\documentclass[journal]{IEEEtran}
\ifCLASSINFOpdf
  \usepackage[pdftex]{graphicx}
\else
  \usepackage[dvips]{graphicx}
\fi
\usepackage{amsmath}
\usepackage{amsfonts}
\usepackage{amssymb}
\usepackage{booktabs}
\usepackage{multirow}
\usepackage{array}
\usepackage{makecell, booktabs}
\usepackage[numbers,sort&compress]{natbib}

\begin{document}
%
\title{RankFormer: A Propose-then-Select Transformer for Multi-Agent Multimodal Trajectory Prediction}

%
%
\author{Diyi~Liu, Zihan~Niu, Tu~Xu, Xingchen~Zhang and~Lishan~Sun 
\thanks{D. Liu, Z. Niu, Xingchen~Zhang and~Lishan~Sun are with the College of Metropolitan Transportation, Beijing University of Technology, Beijing, China. E-mail for the corresponding author: (diyi.liu@bjut.edu.cn). Tu~Xu is with Zhejiang Police College, Zhejiang, China.}
\thanks{Manuscript received xxxx xx, 2026; revised xxxx xx, 2026.}}

\maketitle

\begin{abstract}
Predicting traffic agent trajectories plays an important role in autonomous driving, traffic operations, transportation safety analysis, etc. Although many deep learning algorithms are devised to predict future agent trajectories, the trajectory prediction problem is still challenging due to the complexity of decision-making process, interactions with surrounding vehicles, and the existence of multiple possible intentions for the traveling agents even under similar scenarios. Most existing methods are limited by the requirement of graph structures (e.g., Graph Neural Network) or the requirement of manually labeled intentions. In this study, we propose a pure Transformer-based deep learning model for multi-modal trajectory prediction considering temporal dependencies and agent-agent spatial interactions. After encoding the historical trajectories, two parallel decoders are employed to generate trajectories and probabilities on separate decoder tracks. The model is evaluated on two real-world datasets, one highway dataset and the other pedestrian dataset with solid performance. One important insight is that following a ``propose-then-select'' strategy, the agent-agent spatial interactions are only considered for probability estimation instead of trajectory generation. In summary, the proposed model provides a potential direction to design more robust and effective multi-modal trajectory prediction models.
\end{abstract}

\begin{IEEEkeywords}
Multi-Agent Trajectory Prediction,  Multi-Modal Trajectory Prediction.
\end{IEEEkeywords}

%
\IEEEpeerreviewmaketitle

\section{Introduction}

\IEEEPARstart{T}{raffic} agent trajectory prediction plays an important role in autonomous driving, traffic operations, transportation safety analysis, etc. Despite the development of deep learning models, the agent trajectory prediction task is still challenging due to the complexity of decision-making process, interactions with surrounding agents, and the existence of multiple possible intentions for the traveling agents even under similar scenarios.

During the last decade, the fast development of deep learning models provide a chance to predict trajectories by a data driven approach. Models like Recurrent Neural Networks (RNN), Long-short Term Memory (LSTM), Gated Recurrent Units (GRU) can interpret individual trajectories as a time series to predict future trajectories. To better consider the interactions between agents, multiple approaches are designed to model agent-agent interactions. Social LSTM is one of the early successful work predicting agent trajectories. With the development of the Graph Neural Network (GNN), many studies are using GNN to further model the interactions among agents. Recently, models are shifting to the Transformer architecture to replace traditional RNN/LSTM blocks for both the encoder and decoder modules to achieve better modeling capabilities. Besides the enhancement by Transformer architecture, another benefits is the usage of attention weights to interpret the model's behavior.

Considering the complexity of decision-making process, multiple intentions, or multi-modal capabilities, multiple possible trajectories should be available at the same time for each agent. Thus, another important direction is to design a model to generate multi-modal trajectories. To generate multi-modal trajectories, generative models like Generative Adversarial Networks (GAN), Variational Autoencoders (VAE) are widely used. However, these models tend to learn a latent space to generate similar trajectories without a drastic differences in the pattern. Another limitation is that many of these models lack the ability to decide likelihood of each generated trajectory.

\begin{figure}[!t]
    \centering
    \includegraphics[width=3in]{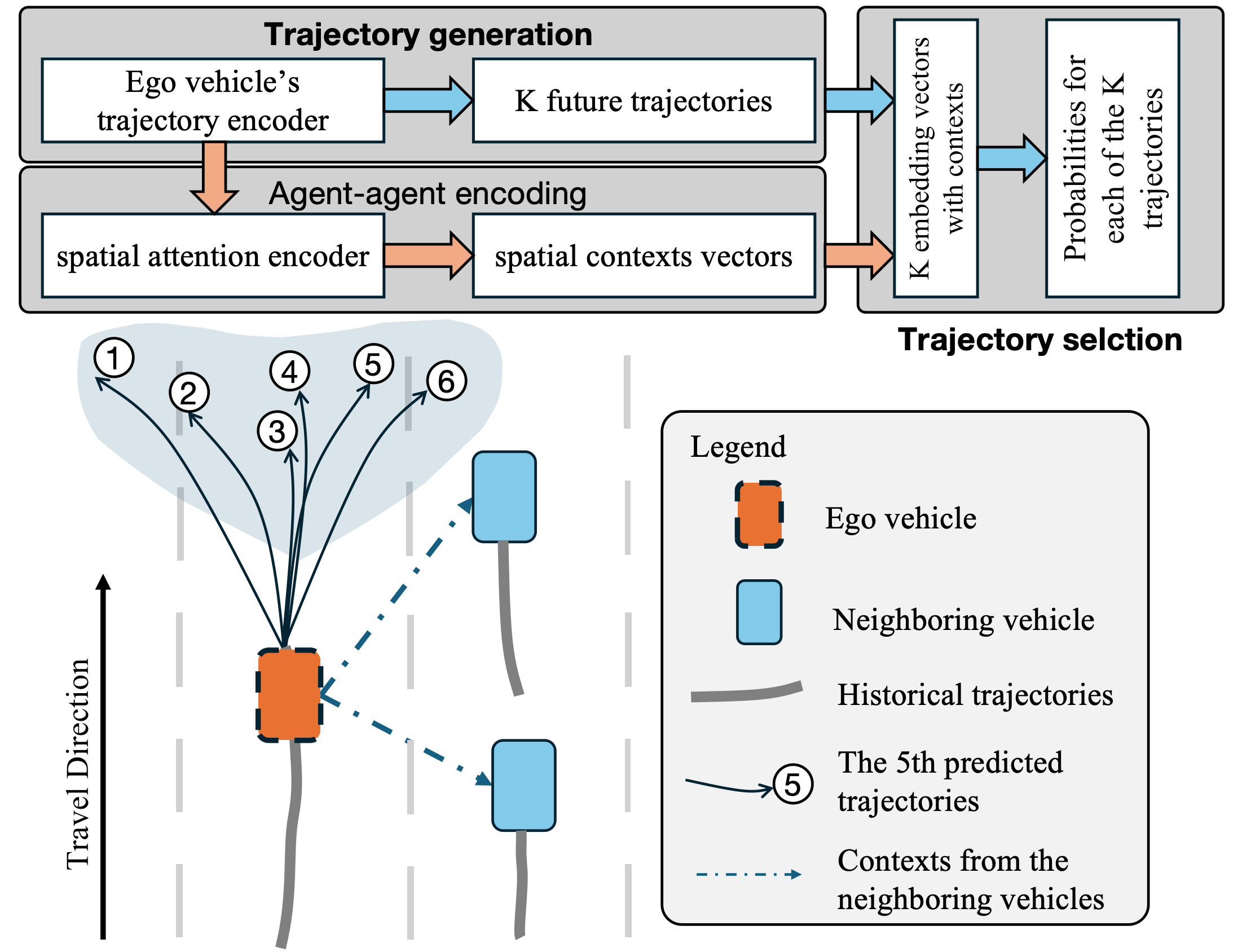}
    \caption{A demonstration of the multi-modal trajectory prediction process. Different from traditional methods, we divide the trajectory prediction process into three steps with two parallel tracks. The first track generates $K$ trajectories corresponding to $K$ modes, the second track generates probabilities for each mode considering spatiotemporal interactions among vehicles/agents as well as considering the inter-modal attentions to estimate probabilities for each path.}
    \label{fig:demo_idea}
\end{figure}

Considering the recent development in agent trajectory prediction, this study proposes a Transformer-based model to generate multi-modal trajectories along with probabilities of each trajectory. Given this requirement, Figure \ref{fig:demo_idea} illustrates the core idea by providing an intuitive summary of the overall roposed strategy. After encoding the historical trajectories, the decoder track generates $K$ trajectories corresponding to $K$ modes. Another decoder generates probabilities for each mode considering spatiotemporal interactions among agents. The final outputs are gathered by selecting the trajectory with the highest probability as the final prediction. In summary, we designed a Transformer-based model named RankFormer, a multi-modal trajectory prediction model with the following contributions:
\begin{itemize}
    \item We apply a ``propose-then-select'' strategy to generate multi-modal trajectories with one decoder before generating the probabilities of each trajectory with another decoder. We hypothesize although the same agent can have many potential trajectories under similar contexts, the impacts of surrounding agents can mainly described as a shifts in the trajectory probabilities.
    \item We make detailed improvements by proposing a new spatial attention module considering inter-agent motions and dynamics and apply an inter-modal attention module to let different trajectory modes communicate to make better probability estimations.
    \item We analyze model outputs quantitatively with ablation studies and compare with some classical methods. Besides, we also apply a series of qualitative analysis by visualizing multi-modal trajectories, agent-agent interactions, cross-attention temporal weights, etc.
\end{itemize}


\section{Related Works}
\subsection{Early Studies}
Vehicle trajectory prediction has been studied for decades and is often organized into physics-based, maneuver-based, and interaction-aware approaches \cite{rudenko2020human}. Early methods extrapolate motion with kinematic assumptions such as constant velocity (CV), constant acceleration (CA), and constant turn-rate-and-acceleration (CTRA) models, typically implemented via Kalman-family filters for state estimation and short-horizon forecasting \cite{schubert2008comparison, blom1988imm}. These models are efficient and interpretable, but they treat future motion as largely deterministic and degrade over longer horizons because driver intent and multi-agent coupling are not explicitly represented. To address this gap, probabilistic maneuver models were introduced: Hidden Markov Models (HMMs) learn transition structure from historical trajectories \cite{ye2016vehicle}, while Bayesian networks combine vehicle dynamics with map context to infer discrete intentions such as lane changes and turns before generating associated paths \cite{lefevre2011exploiting, kasper2012object}. Interaction effects were also modeled explicitly before the deep-learning era, for example through social-force formulations that treat nearby agents as repulsive/attractive influences on motion \cite{helbing1995social}, and through early multi-agent trajectory studies in crowded scenes \cite{pellegrini2010improving}. As large-scale trajectory data became available, researchers moved from hand-crafted extrapolation toward data-driven characterization of route-level movement patterns via trajectory clustering and classification \cite{kim_spatial_2015}, motivating the learning-based encoders discussed in the next subsection.

\subsection{Information Encoding}
Deep learning predictors encode agent motion from historical trajectories and neighboring context. For temporal encoding, RNNs/LSTMs/GRUs and TCN/CNN variants are widely used; attention-based encoders further improve dependency modeling and explainability in highway and urban traffic \cite{messaoud_attention_2021, lin_vehicle_2022, xu_leveraging_2023, jiang_intention-aware_2023}.

Although often described as interaction modeling, most multi-agent modules function as spatial encoders that aggregate neighboring agents into the target representation. Social LSTM introduced pooling grids for this purpose \cite{alahi_social_2016}; GNN-based methods such as GRIP, VectorNet, and GAT variants represent agents as nodes and relations as edges \cite{li2019grip, gao2020vectornet, zhang_trajectory_2022, gao_surrounding_2024, yang_multi-task_2024}. Similar ideas extend to mixed traffic \cite{liu2022hega, zhou2022hivt, lai2020trajectory, chandra2019traphic, ma2019trafficpredict} and socially aware representation learning \cite{liu2021social}. Predefined graph topology can nevertheless limit long-range or heterogeneous coupling, motivating attention-based spatial encoding in later Transformer predictors. Once temporal and spatial context are encoded, the next challenge is to generate and score multiple plausible futures, as discussed below.

\subsection{Multi-modality Prediction}
Travel trajectories, by nature, are a stochastic process over a range of future trajectories with different intentions. Thus, it is important to generate multiple trajectories given the same historical condition. Early generative approaches include Social-GAN \cite{gupta2018social}, Sophie \cite{sadeghian2019sophie}, and graph-attention variants such as Social-BIGAT \cite{kosaraju2019social}. Latent-variable and diffusion models such as Trajectron++ \cite{salzmann2020trajectron++}, Leapfrog Diffusion \cite{mao2023leapfrog}, and MotionDiffuser \cite{jiang2023motiondiffuser} further model trajectory generation as conditional sampling or iterative denoising. These methods focus primarily on mode \emph{generation} without much probability estimation; scoring and selecting a single best mode is discussed in the next subsection.

\subsection{Mode Selection and Probability Estimation}
Multimodal predictors must also identify the most likely mode for top-1 deployment. WTA training is standard for this setting: the model outputs $K$ candidates with associated probabilities, but only the best-matching trajectory receives regression loss while all modes receive classification supervision \cite{cui2019multimodal}. Prior work embeds probability estimation in anchor classifiers \cite{chiu2019multipath, phanminh2020covernet}, goal-conditioned decoders \cite{mangalam2020pecnet}, or unified Transformer heads \cite{yuan2021agentformer, shi2022mtr}. RankFormer adopts a propose-then-select design: one decoder proposes $K$ label-free modes using sampled WTA regression, and a separate Path Ranker learns probabilities from distance-based soft targets using interaction-aware context.

\subsection{Model Interpretation}
Beyond accurate multi-modal forecasting, interpretability is increasingly important for safety-critical deployment and downstream planning. Early deep learning predictors addressed this through intrinsic attention mechanisms: spatial--temporal attention in LSTM encoders \cite{messaoud_attention_2021, lin_vehicle_2022}, inter-vehicle and intention-level attention in Transformer models \cite{jiang_intention-aware_2023}, and temporal-pattern attention with layer-wise mutual-information analysis \cite{hu_trajectory_2023}. These methods visualize which neighboring agents and historical time steps receive higher weight, and case studies on lane changes or dense traffic often reveal patterns consistent with human driving behavior.

However, attention-based visualization has important limitations. Attention weights alone do not necessarily reflect causal contribution to the output \cite{jain2019attention}, and attribution analyses indicate that many predictors rely predominantly on the target agent's own history even when interaction modules are present \cite{makansi2021you}. Layer-wise studies further show that salient driving cues often reside in intermediate features rather than raw attention maps \cite{hu_trajectory_2023}. These findings support RankFormer's propose-then-select design: $K$ trajectories are proposed from encoded history, while spatial attention in the Path Ranker captures how neighbors shift mode probabilities.

\begin{figure*}[!t]
    \centering
    \includegraphics[width=\textwidth]{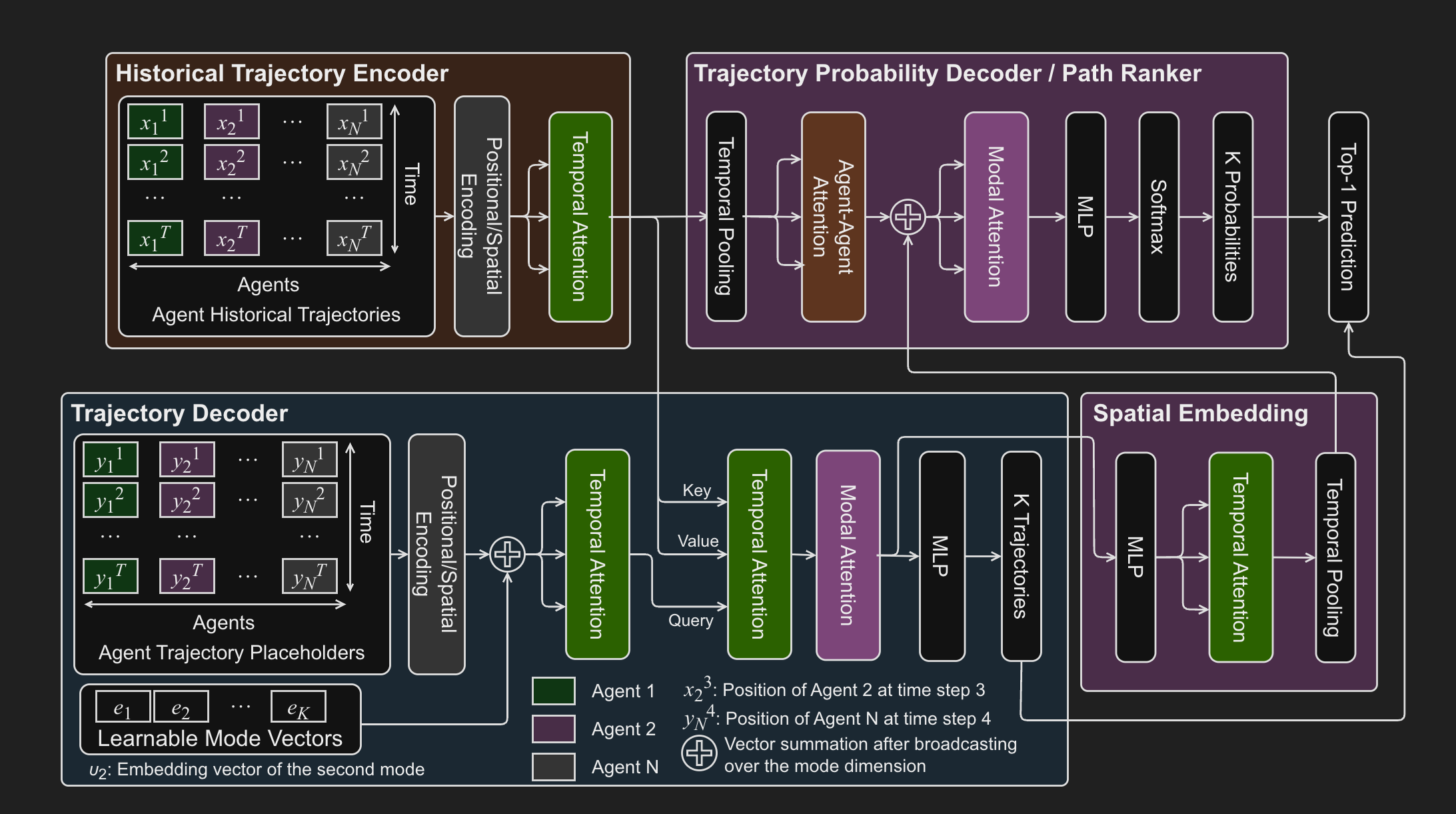}
    \caption{An Overview of the proposed model framework with one trajectory encoder and two parallel decoders to generate multi-modal trajectories and probabilities. For inference, the final prediction is selected based on the highest probability among all modes. During training, the total objective combines sampled WTA trajectory regression and a soft cross-entropy loss for probability selection. A spatial embedding layer }
    \label{fig:overview}
\end{figure*}

\section{Methodology}

\subsection{Problem Definition}
Given time $t$, the motion state of a vehicle $i$ is described by a tuple $s_i^t = (c_i^t, v_i^t, a_i^t, \theta_i^t, y_i^t)$, where $c_i^t$, $v_i^t$, $a_i^t$, $\theta_i^t$ and $y_i^t$ respectively denote coordinates, velocity, acceleration, angle, and yaw. Given historical time period indexed by $t \in \{1, \ldots , T_{hist}\}$, the history trajectories are represented as a series of coordinates: 

\begin{equation}
    \begin{aligned}
        \mathbf{X} & = \{ c_i^{1:T_{hist}} , i \in [1, N] \} \\
        \mathbf{S} & = \{ s_i^{1:T_{hist}} , i \in [1, N] \}
    \end{aligned}
\end{equation}

Note that given a series of $c_i^t$ with high time resolutions (e.g., 10 Hz), all other features above are derivable. Depending on the application, one can use $\mathbf{X}$ or $\mathbf{S}$ to encode the history trajectories. In other words, one can apply $S = \text{f} (X)$ to derive the other features from the coordinates.

To predict the future trajectory during the next $T_{pred}$ timestamps ranging from $(T_{hist}+1)$ to $(T_{hist} + T_{pred})$, the future trajectories are represented as:

\begin{equation}
    \mathbf{Y} = \{ c_i^{T_{hist}+1:T_{hist}+T_{pred}} , i \in [1, N] \}
\end{equation}

For multi-modal prediction, there are $K$ different modal trajectories indexed by $k \in \{1, \ldots, K\}$ represented as follows:

\begin{equation}
    \mathbf{\hat{Y}}_{i,k} = \{ c_{i,k}^{T_{hist}+1:T_{hist}+T_{pred}} , k \in [1, K]\}
\end{equation}

where each $c_{i,k}$ is the $k$ th mode of estimated trajectory for agent $i$. The model also outputs a predicted probability $\hat{P}_{i,k}$ for each mode, with $\hat{\mathbf{P}}_i = (\hat{P}_{i,1}, \ldots, \hat{P}_{i,K})$ and $\sum_{k=1}^{K} \hat{P}_{i,k} = 1$ for each agent $i$.

During inference, for each agent $i$, we select the mode with the highest predicted probability and use its trajectory as the final prediction:

\begin{equation}
    w_i^{\mathrm{pred}} = \arg\max_{k \in \{1, \ldots, K\}} \hat{P}_{i,k},
    \qquad
    \forall i \in \{1, \ldots, N\}
\end{equation}

\begin{equation}
    \hat{\mathbf{Y}}_i^{\mathrm{pred}}
    = \hat{\mathbf{Y}}_{i, w_i^{\mathrm{pred}}}.
\end{equation}

\subsection{Overall Architecture}
Figure \ref{fig:overview} gives an overview of the proposed architecture of the RankFormer. The architecture is composed of one Trajectory Encoder and two parallel decoders. The Trajectory Encoder is used to encode the historical trajectories. The first decoder, Trajectory Decoder, is to generate multi-modal trajectories. The second decorder, the PathRanker, is to estimate the probabilities of each generated trajectory. Several Transformer blocks with similar details are devised to apply attention mechanism over different dimensions. For example, the Temporal and Spatial Attention modules communicate information along time and space dimensions, respectively. An Inter-modal Attention module communicates information among a modal dimension across different modes. By observing Figure \ref{fig:overview}, the interactions between different vehicles (i.e., Spatial Attention module) are modeled solely to shift the probabiliry of each trajectory. Finally, combined with $K$ trajectories and their probabilities, the final prediction is selected based on the highest probability among all modes. The remainder of the section will detail the architecture of important modules shown in Figure \ref{fig:overview}.

\subsection{Positional and Spatial Encoding}
For Trajectory Encoder, temporal information is encoded using standard Positional Encoding:

\begin{equation}
\begin{aligned}
PE_{(t, 2i)} &= \sin\!\left( \frac{t}{10000^{2i / d}} \right) \\
PE_{(t, 2i+1)} &= \cos\!\left( \frac{t}{10000^{2i / d}} \right)
\end{aligned}
\end{equation}

where $t \in [1, \ldots, T_{hist} + T_{pred}]$ is anytime during the historical and future time frames. For Spatial Encoding, the historical trajectories are encoded using a Multi-Layer Perceptron (MLP):

\begin{equation}
\begin{aligned}
SE(\mathbf{X}) = \text{MLP} ( \text{f} (\mathbf{X}) )
\end{aligned}
\end{equation}

where $f(\cdot)$ is the (optional) function to augment the coordinates to include the other features. For the Trajectory Decoder, to distinguish between $K$ trajectories, the encoding becomes:

\begin{equation}
\begin{aligned}
SE(\mathbf{X}) = \text{MLP} ( \text{f} (\mathbf{X}) ) + \mathbf{e}_{\text{mode}}
\end{aligned}
\end{equation}

where $\mathbf{e}_{mode}$ is a learnable vector for the $k$ th mode. In summary, the spatial and temporal encoding vectors are combined to generate the input feature for the Temporal Attention Module as:

\begin{equation}
\begin{aligned}
\mathbf{X}_{enc} = \text{SE}(\mathbf{X}) + \text{PE}(\mathbf{X})
\end{aligned}
\end{equation}

\subsection{Temporal Attention Module}
Let $\mathbf{X}_n^{(l)}\in \mathbb{R}^{T \times d}$ denote the input features of agent $n \in \{1, \ldots, N\}$ at Transformer layer $l \in \{0, \ldots, L-1\}$, where $T$ is the number of time steps and $d$ is the hidden dimension. At layer $l$, $\mathbf{X}_n^{(l)}$ is projected into $H$ attention heads to obtain query $\mathbf{Q}^{(h)}_n$, key $\mathbf{K}^{(h)}_n$, and value $\mathbf{V}^{(h)}_n$ matrices:

\begin{equation}
    \begin{aligned}
    \mathbf{Q}^{(h)}_n & = \mathbf{X}_n^{(l)} W_q^{(h)} \\
    \mathbf{K}^{(h)}_n & = \mathbf{X}_n^{(l)} W_k^{(h)} \\
    \mathbf{V}^{(h)}_n & = \mathbf{X}_n^{(l)} W_v^{(h)} 
    \end{aligned}
\end{equation}

Given the query, key, and value matrices, a temporal attention mechanism is used to incorporate the dynamics of each agent: 

\begin{equation}
\text{head}_h = \text{softmax}\!\left( \frac{\mathbf{Q}^{(h)}_n\mathbf{K}_n^{(h)\top}}{\sqrt{d_k}}\right) \mathbf{V}_n^{(h)}
\end{equation}

Concatenating $H$ heads, the multi-head attention output becomes:

\begin{equation}
\text{MHA}(\mathbf{X}_n^{(l)}) = \text{Concat}(\text{head}_1, \dots, \text{head}_H) \, \mathbf{W}_O
\end{equation}

Following the pre-LayerNorm residual pattern, the full Temporal Attention block updates the layer input as:

\begin{equation}
\begin{aligned}
\tilde{\mathbf{X}}_n^{(l)} &= \mathbf{X}_n^{(l)} + \text{MHA}\big(\text{LayerNorm}(\mathbf{X}_n^{(l)})\big) \\
\mathbf{X}_n^{(l+1)} &= \tilde{\mathbf{X}}_n^{(l)} + \text{FFN}\big(\text{LayerNorm}(\tilde{\mathbf{X}}_n^{(l)})\big)
\end{aligned}
\end{equation}

where $\mathbf{X}_n^{(l+1)}$ is the output of layer $l$ and serves as the input to layer $l+1$. Each Transformer block's FFN is implemented as a two-layer MLP. Both encoder and decoder use the same Temporal Attention Module. However, the decoder applies the module in parallel $K$ times to incorporate $K$ different intentions.

\subsection{Agent-Agent Attention Module}
The spatial attention module follows the same layer-indexed residual structure as the Temporal Attention module, except that the attention network communicates information among vehicles rather than across time frames. Let $\mathbf{X}_{(t)}^{(l)}\in \mathbb{R}^{N \times d}$ denote the input features of all $N$ vehicles at time step $t \in \{1, \ldots, T_{\mathrm{hist}} + T_{\mathrm{pred}}\}$ and layer $l$. At each time step, the agent features are projected into $H$ attention heads to obtain query $\mathbf{Q}^{(h)}_t$, key $\mathbf{K}^{(h)}_t$, and value $\mathbf{V}^{(h)}_t$ matrices:

\begin{equation}
    \begin{aligned}
    \mathbf{Q}^{(h)}_t & = \mathbf{X}_{(t)}^{(l)} W_q^{(h)} \\
    \mathbf{K}^{(h)}_t & = \mathbf{X}_{(t)}^{(l)} W_k^{(h)} \\
    \mathbf{V}^{(h)}_t & = \mathbf{X}_{(t)}^{(l)} W_v^{(h)}
    \end{aligned}
\end{equation}

The spatial attention head is then computed as:

\begin{equation}
\text{head}_h = \text{softmax}\!\left( \frac{\mathbf{Q}^{(h)}_t\mathbf{K}_t^{(h)\top}}{\sqrt{d_k}} + \mathbf{B}^{(h)} \right) \mathbf{V}_t^{(h)}
\end{equation}

where $\mathbf{B} \in \mathbb{R}^{N \times N}$ denotes the relative bias matrix to incorporate inter-vehicle motions. Specifically, $b_{ij}$, an value in matrix at $i$ th row and $j$ th column of $\mathbf{B}$, denotes the relative bias between agent $i$ and $j$, respectively. $b_{ij}$ is calculated by fusing relative information over a MLP fusing relative difference in historical trajectories:
\begin{equation}
    b_{ij} = \text{MLP}([\Delta{s}_{ij}])
\end{equation}

where $\Delta{s_{ij}} = s_i - s_j$ denote the difference between agent $i$ and $j$'s motion features (e.g., speed, yaw). Instead of using normalized distance, differentiation is applied to incorporate asymmetrical relationships between two vehicles. The multi-head output, feed-forward sublayer, and pre-LayerNorm residual update follow the same form as in the Temporal Attention Module, producing $\mathbf{X}_{(t)}^{(l+1)}$ from $\mathbf{X}_{(t)}^{(l)}$ at each time step.

\subsection{Modal Attention Module}
The Modal Attention module follows the same layer-indexed residual structure as the Temporal Attention module, except that the attention network communicates information among $K$ trajectory modes instead of across time frames or vehicles. Before entering the mode probability decoder, encoder outputs are pooled over the time dimension to obtain a compact representation for each vehicle and intention mode. Let $\mathbf{Z}_n^{(l)} \in \mathbb{R}^{K \times d}$ denote the pooled modal features of all $K$ intention modes for vehicle $n \in \{1, \ldots, N\}$ at layer $l \in \{0, \ldots, L-1\}$, stacked as $\mathbf{Z}^{(l)} \in \mathbb{R}^{N \times K \times d}$ across vehicles. At layer $l$, $\mathbf{Z}_n^{(l)}$ is projected into $H$ attention heads to obtain query $\mathbf{Q}^{(h,\mathrm{m})}_n$, key $\mathbf{K}^{(h,\mathrm{m})}_n$, and value $\mathbf{V}^{(h,\mathrm{m})}_n$ matrices:

\begin{equation}
    \begin{aligned}
    \mathbf{Q}^{(h,\mathrm{m})}_n & = \mathbf{Z}_n^{(l)} W_q^{(h,\mathrm{m})} \\
    \mathbf{K}^{(h,\mathrm{m})}_n & = \mathbf{Z}_n^{(l)} W_k^{(h,\mathrm{m})} \\
    \mathbf{V}^{(h,\mathrm{m})}_n & = \mathbf{Z}_n^{(l)} W_v^{(h,\mathrm{m})}
    \end{aligned}
\end{equation}

Given the query, key, and value matrices, a modal attention mechanism is used to incorporate interactions among the $K$ candidate intentions:

\begin{equation}
\text{head}_h^{\mathrm{m}} = \text{softmax}\!\left( \frac{\mathbf{Q}^{(h,\mathrm{m})}_n\mathbf{K}_n^{(h,\mathrm{m})\top}}{\sqrt{d_k}} \right) \mathbf{V}_n^{(h,\mathrm{m})}
\end{equation}

Concatenating $H$ heads, the multi-head attention output becomes:

\begin{equation}
\text{MHA}^{\mathrm{m}}(\mathbf{Z}_n^{(l)}) = \text{Concat}(\text{head}_1^{\mathrm{m}}, \dots, \text{head}_H^{\mathrm{m}}) \, \mathbf{W}_O^{\mathrm{m}}
\end{equation}

Following the pre-LayerNorm residual pattern, the full Modal Attention block updates the layer input as:

\begin{equation}
\begin{aligned}
\tilde{\mathbf{Z}}_n^{(l)} &= \mathbf{Z}_n^{(l)} + \text{MHA}^{\mathrm{m}}\big(\text{LayerNorm}(\mathbf{Z}_n^{(l)})\big) \\
\mathbf{Z}_n^{(l+1)} &= \tilde{\mathbf{Z}}_n^{(l)} + \text{FFN}\big(\text{LayerNorm}(\tilde{\mathbf{Z}}_n^{(l)})\big)
\end{aligned}
\end{equation}

where $\mathbf{Z}_n^{(l+1)}$ is the output of layer $l$ and serves as the input to layer $l+1$. The Modal Attention module is applied as a part of the Trajectory Probability Decoder (Path Ranker), where $K$ parallel modes are first proposed on pooled encoder features and then refined through cross-modal interaction before probability estimation.

\subsection{Path Ranker: Prediction Heads}
Following the propose-then-select strategy in Figure~\ref{fig:overview},
the Trajectory Decoder proposes $K$ candidate trajectories for each agent. A second decoder named the Path Ranker scores them using interaction-aware context from both the Trajectory Encoder and Trajectory Decoder. Let $\mathbf{H}_n^{\mathrm{enc}}\in\mathbb{R}^{T\times d}$
denote the encoded features of agent~$n$, and let $\mathbf{\hat{X}}_n^{(-1)}\in\mathbb{R}^{K\times T_{\mathrm{pred}}\times d}$ denote the mode-conditioned decoder features after the second temporal attention block (which attends to encoder queries). $-1$ denotes value is the output of the second as well as the last temporal attention block.

\paragraph{Trajectory head}
Candidate trajectories are decoded from the mode-conditioned features using a two-layer MLP as trajectory projection head:
\begin{equation}
    \hat{\mathbf{Y}}_n
    = \mathrm{MLP}\big(\mathrm{MLP}(\mathbf{\hat{X}}_n^{(-1)})\big)
    \in \mathbb{R}^{K\times T_{\mathrm{pred}}\times 2}
\end{equation}

\paragraph{Context branch}
On the Path Ranker side, encoder features first exchange information
across agents and then across time:
\begin{equation}
\begin{aligned}
    \mathbf{C}_n
    &=
    \mathrm{TemporalAttn}\!\big(
        \mathrm{SpatialAttn}(\mathbf{H}^{\mathrm{enc}})
    \big)_n
\end{aligned}
\end{equation}
where $\mathbf{H}^{\mathrm{enc}}$ is the encoded features of all agents, and $\mathbf{C}_n$ is the context features of agent $n$. $\mathbf{C}_n\in\mathbb{R}^{T_{\mathrm{pred}}\times d}$ denotes the context features of agent $n$. The resulting context is pooled over time (e.g., by taking average over the time dimension) to a vector $\bar{\mathbf{c}}_n\in\mathbb{R}^{d}$.

\paragraph{Mode fusion and ranking}
Trajectory Decoder mode features are likewise pooled over the prediction horizon
to $\bar{\mathbf{X}}_n\in\mathbb{R}^{K\times d}$. Context and mode
embeddings are fused along the mode axis by broadcasting context and
adding the learnable intention embeddings $\mathbf{e}_k$:
\begin{equation}
    \mathbf{Z}_n
    =
    \bar{\mathbf{X}}_n
    +
    \mathbf{1}_K\,\bar{\mathbf{c}}_n^{\top}
\end{equation}
which matches the $\oplus$ operation in Figure~\ref{fig:overview} (``modal over modal dimension''). An inter-modal attention module then ``ranks'' the $K$ candidates by a Modal Attention layer before a scoring head:
\begin{equation}
    \begin{aligned}
    \mathbf{Z}_n' & = \mathrm{ModalAttn}(\mathbf{Z}_n), \\
    \hat{P}_n & = \mathrm{Softmax}\!\big( \mathrm{MLP}\big(\mathrm{MLP}(\mathbf{Z}_n')\big) \big)
    \in \Delta^{K-1}.
    \end{aligned}
\end{equation}

\subsection{Coordinate Normalization}
\label{sec:normalization}

The model operates on normalized motion features, with different coordinate transformations for the pedestrian and vehicle datasets. The same transformation is applied to observed and target positions within each sample. Displacement errors are converted back to meters for evaluation.

\paragraph{Pedestrian scene coordinates}
For each scene window, let $\mathcal{H}$ denote the agents with at least one valid historical observation, and let $t_i^{\mathrm{last}}$ be agent $i$'s last valid observation time. We define a shared scene origin using only the observed history:
\begin{equation}
    \mathbf{o}
    = \frac{1}{|\mathcal{H}|}
      \sum_{i\in\mathcal{H}} \mathbf{x}_{i,t_i^{\mathrm{last}}}.
\end{equation}
Observed and future positions are transformed as
\begin{equation}
    \tilde{\mathbf{x}}_{i,t}
    = \frac{\mathbf{x}_{i,t}-\mathbf{o}}{s},
    \qquad
    \tilde{\mathbf{y}}_{i,t}
    = \frac{\mathbf{y}_{i,t}-\mathbf{o}}{s},
\end{equation}
where $s>0$ is a fixed physical length scale shared across scenes and both coordinate axes. The origin remains fixed throughout the window, and all agents use the same origin, preserving their relative positions up to the common scale. The single-agent variant uses the original scene's origin before separating its agents into individual samples. No future labels enter the origin calculation, and no dataset-level means or standard deviations are fitted.

Speed, velocity components, and acceleration are divided by $s$ without subtracting a population mean; time remains measured in seconds. Coordinates are not clipped to a fixed range, and masks retain the distinction between valid observations and padding. Predicted positions are restored to world coordinates by $\hat{\mathbf{y}}_{i,k,t}=s\hat{\tilde{\mathbf{y}}}_{i,k,t}+\mathbf{o}$. Consequently, normalized Euclidean position error equals error in meters divided by $s$, giving both axes equal physical weight.

\paragraph{Vehicle standardization}
For the vehicle dataset, each motion-feature channel $c$ is standardized using its training-set mean $\mu_c$ and standard deviation $\sigma_c$:
\begin{equation}
    \tilde{z}_c = \frac{z_c-\mu_c}{\sigma_c}.
\end{equation}
These statistics are computed over valid entries in the training split and reused for validation and testing. Position predictions are converted back using the corresponding coordinate statistics. Because the two position channels may have different standard deviations, Euclidean error in normalized vehicle coordinates weights the two axes differently from error in meters; it is not generally related to meter error by a single scale factor.

\subsection{Objective function \& model evaluation}

We describe the objective for point-valued trajectory outputs. At inference, the Path Ranker selects the highest-probability proposal, $w_i^{\mathrm{pred}} = \arg\max_k \hat{P}_{i,k}$. During training, no discrete intention labels are required: trajectory regression uses a sampled geometric assignment, while the ranker learns a separate soft distribution derived from the errors of all $K$ proposals.

\paragraph{Masked per-mode displacement error}
Let $m_{i,t}\in\{0,1\}$ indicate a valid future label, and let $\mathcal{S}$ contain the valid agents with at least one such label. For $i\in\mathcal{S}$ and mode $k$, the regression error is the average Euclidean displacement over valid future steps in the normalized coordinates of Section~\ref{sec:normalization}:
\begin{equation}
    d_{i,k}
    = \frac{\sum_{t=1}^{T_{pred}} m_{i,t}
      \bigl\|\hat{\tilde{\mathbf{y}}}_{i,k,t}
                 -\tilde{\mathbf{y}}_{i,t}\bigr\|_2}
      {\sum_{t=1}^{T_{pred}} m_{i,t}}.
\end{equation}
Thus, the per-mode loss is normalized ADE rather than MSE. The geometric winner is
\begin{equation}
    w_i^{\mathrm{geo}} = \arg\min_{k\in\{1,\ldots,K\}} d_{i,k}.
\end{equation}

\paragraph{Sampled trajectory regression}
To give initially inaccurate modes opportunities to learn, we use $\varepsilon$-greedy Winner-Takes-All (WTA) assignment. For each agent, one regression mode $a_i$ is sampled according to
\begin{equation}
    \Pr(a_i=k)
    = (1-\varepsilon)\,\mathbf{1}[k=w_i^{\mathrm{geo}}]
      + \frac{\varepsilon}{K}.
\end{equation}
The trajectory loss averages the selected errors over supervised agents:
\begin{equation}
    \mathcal{L}_{\mathrm{traj}}
    = \frac{1}{|\mathcal{S}|}\sum_{i\in\mathcal{S}} d_{i,a_i}.
\end{equation}
Only the selected proposal contributes a direct regression term for each agent in an update. Exploration changes which proposal is selected; it does not apply a weighted regression loss to all proposals simultaneously. The exploration rate is annealed to zero as specified in Section~\ref{sec:experiment}, recovering hard WTA. Validation loss is evaluated without random exploration.

\paragraph{Distance-soft ranker supervision}
The ranker's target distribution is computed independently of the sampled regression mode:
\begin{equation}
    q_{i,k}
    = \frac{\exp\!\left(-\operatorname{sg}(d_{i,k})/\tau\right)}
      {\sum_{j=1}^{K}\exp\!\left(-\operatorname{sg}(d_{i,j})/\tau\right)},
\end{equation}
where $\tau>0$ is a temperature in normalized ADE units and $\operatorname{sg}$ denotes stop-gradient. Thus, the target construction supplies no gradient through $d_{i,k}$. Similar-quality proposals share target probability, while smaller $\tau$ concentrates supervision more strongly on the lowest-error modes. These targets express relative geometric fit rather than calibrated probabilities of future events.

The Path Ranker is trained with soft cross-entropy:
\begin{equation}
    \mathcal{L}_{\mathrm{prob}}
    = -\frac{1}{|\mathcal{S}|}
      \sum_{i\in\mathcal{S}}\sum_{k=1}^{K}
      q_{i,k}\,\log \hat{P}_{i,k}.
\end{equation}
The total objective is
\begin{equation}
    \mathcal{L} = \mathcal{L}_{\mathrm{traj}} + \alpha \cdot \mathcal{L}_{\mathrm{prob}},
\end{equation}
where $\alpha\geq0$ balances trajectory regression and mode ranking. Agents without valid future labels contribute to neither loss, although their observed histories may remain in the interaction context; batches without supervised agents skip the optimizer update. Normalization affects both the magnitude of $\mathcal{L}_{\mathrm{traj}}$ and the physical interpretation of $\tau$. For pedestrian coordinates with scale $s$, temperature $\tau$ corresponds to a distance scale of $s\tau$ meters.

\paragraph{Evaluation metrics}
We report Average Displacement Error (ADE) and Final Displacement Error (FDE) in meters. For multi-modal prediction with $K$ modes, we report \emph{top-1} ADE/FDE using the highest-probability mode ($w_i^{\mathrm{pred}}$), which matches the inference-time selection rule. We also report oracle $\mathrm{minADE}_K$ and $\mathrm{minFDE}_K$ by independently minimizing each metric over the $K$ proposals for each agent. These choices need not coincide with each other or with the normalized-ADE winner used for training.
\begin{equation}
\mathrm{ADE}
=
\frac{1}{N}\sum_{n=1}^{N}
\frac{1}{T}\sum_{t=1}^{T}
\bigl\| \hat{\mathbf{y}}_{n,t} - \mathbf{y}_{n,t} \bigr\|_2
\end{equation}
\begin{equation}
\mathrm{FDE}
=
\frac{1}{N}\sum_{n=1}^{N}
\bigl\| \hat{\mathbf{y}}_{n,T} - \mathbf{y}_{n,T} \bigr\|_2
\end{equation}
\begin{equation}
\label{eq:minade}
\mathrm{minADE}_K
=
\frac{1}{N}\sum_{n=1}^{N}
\min_{k\in\{1,\ldots,K\}}
\frac{1}{T}\sum_{t=1}^{T}
\bigl\| \hat{\mathbf{y}}_{n,t}^{(k)} - \mathbf{y}_{n,t} \bigr\|_2
\end{equation}
\begin{equation}
\label{eq:minfde}
\mathrm{minFDE}_K
=
\frac{1}{N}\sum_{n=1}^{N}
\min_{k\in\{1,\ldots,K\}}
\bigl\| \hat{\mathbf{y}}_{n,T}^{(k)} - \mathbf{y}_{n,T} \bigr\|_2
\end{equation}

\section{Experiment}
\label{sec:experiment}
Two datasets, one pedestrian dataset (i.e., the ETH dataset \cite{pellegrini2010improving}) and a vehicle trajectory dataset (i.e., the Ubiquitous Traffic Eyes dataset \cite{feng2026ubiquitous}), are used to evaluate the model's performance. The classic ETH dataset contains annotated pedestrian trajectories in 5 different environments (i.e., ETH, Hotel, Univ, Zara1, and Zara2). Each environment has a static camera to capture top view videos to process trajectories at 2.5 fps. Similar to previous studies, 8 timestamps (3.2 seconds) are used for historical information and the following 12 timestamps (4.8 seconds) are used for prediction.

The Ubiquitous Traffic Eyes dataset is used to evaluate the model's performance on vehicle trajectories. Three datasets (i.e., ``SQM-W'', ``SQM-W'', and ``SQM-N\_up'') are used for training. The ``SQM-N\_down'' dataset is selected for testing. This elevated highway is a special merging/splitting zone with a lot of traffic lanes and complex conflict zones. Unlike the old ETH dataset, the whole dataset is processed at 30 fps. In this study, the trajectories are processed at 10 fps. 5 seconds of historical and 5 seconds of future trajectories are used to evaluate the model's effectiveness.

\subsection{Data Processing and Model Configuration}

Each dataset contains vehicle trajectories for tens of minutes. We first chunk the dataset into isolated 10-second chucks. For each chuck, the first 5 seconds are used for historical information and the remaining 5 seconds are used for ground-truth data for prediction. A boolean mask is provided for vehicles entered or left early during the 10-second phase. Thus, all unknown vehicle coordinates are filtered out for both model inputs and outputs.

By default, the model predicts $K=5$ modes of intentions. All Transformer layers have four parallel heads, and each head has a vector with dimension size of 16. A dropout rate of 0.1 is applied to the input and output of the Transformer layers. All models are trained over the dataset for 100 epochs. For trajectory regression, the exploration schedule is $\varepsilon_e=\max\{0,1-(e-1)/10\}$, with epochs indexed from $e=1$. It starts at one in epoch 1 and reaches zero in epoch 11. The default soft-target temperature is $\tau=0.1$ and the ranker loss weight is $\alpha=0.01$. Pedestrian coordinates use a fixed scale of $s=10$ meters per coordinate unit, so this temperature corresponds to 1 meter; vehicle coordinates use the training-set statistics described in Section~\ref{sec:normalization}. Furthermore, the model is trained applying an Adam optimizer with a learning rate of 0.001. Since there are different number of vehicles/pedestrians at different time steps in each scene, the model is simply trained scene by scene.

\subsection{Performance Comparison}
The model performance is reported as ADE/FDE in meters under a 5-second prediction horizon. An ablation study is applied to evaluate the effectiveness of the proposed architecture design. Table~\ref{tab:ablation_vehicle} summarizes an ablation study over spatial vehicle interactions and the number of predicted modes~$K$. For $K=5$, top-1 metrics reflect end-to-end mode selection, whereas oracle metrics report $\mathrm{minADE}_5$/$\mathrm{minFDE}_5$ and quantify the geometric quality of the hypothesis set. Spatial interactions improve both top-1 and oracle performance; with interactions enabled, top-1 ADE/FDE decreases from 2.50/4.88~m to 2.28/4.58~m and oracle ADE/FDE from 1.85/3.37~m to 1.64/3.02~m.

\begin{table}[!h]
    \caption{Ablation studies on the Ubiquitous Traffic Eyes (SQM) vehicle trajectory dataset.
    For $K=5$, \emph{top-1} uses the highest-probability mode; \emph{oracle} reports $\mathrm{minADE}_5$/$\mathrm{minFDE}_5$ (Eqs.~\eqref{eq:minade}--\eqref{eq:minfde}).}
    \label{tab:ablation_vehicle}
    \centering
    \footnotesize
    \renewcommand{\arraystretch}{1.15}
    \setlength{\tabcolsep}{3.5pt}
    \begin{tabular}{l c c l c c}
    \toprule
    Model & \makecell{Spatial\\int.} & $K$ & Selection & ADE $\downarrow$ (m) & FDE $\downarrow$ (m) \\
    \midrule
    Ours & no  & 1 & single mode    & 3.70 & 5.49 \\
    \midrule
    Ours & no  & 5 & top-1          & 2.50 & 4.88 \\
    Ours & no  & 5 & oracle (min@$K$) & 1.85 & 3.37 \\
    \midrule
    Ours & yes & 5 & top-1          & \textbf{2.28} & \textbf{4.58} \\
    Ours & yes & 5 & oracle (min@$K$) & \textbf{1.64} & \textbf{3.02} \\
    \bottomrule
    \end{tabular}
\end{table}

\begin{figure}[!b]
    \centering
    \includegraphics[width=\linewidth]{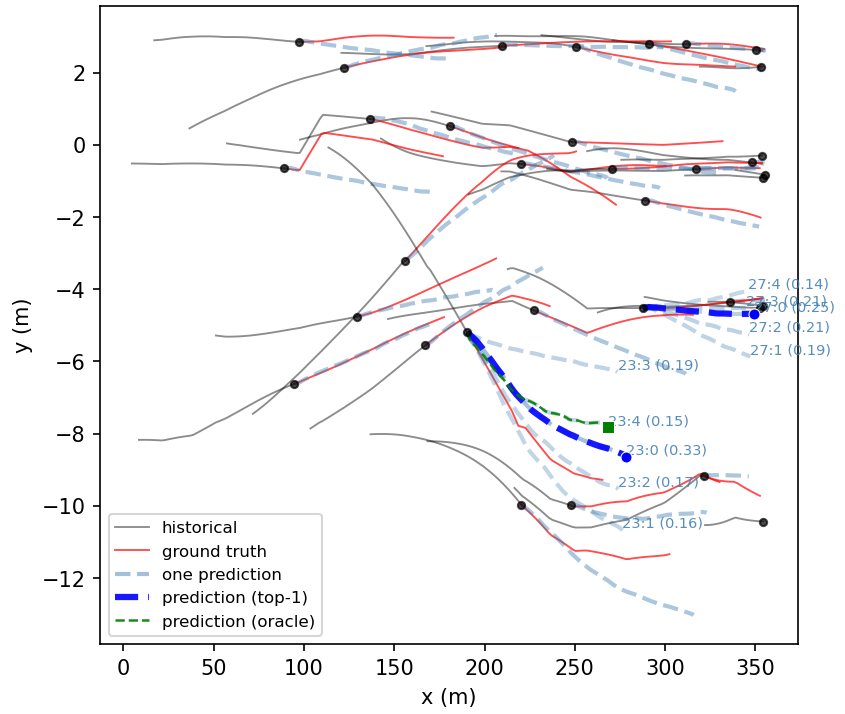}
    \caption{Model predictions vs. ground truth trajectories for the vehicle trajectory dataset. Two typical vehicles are selected to visualize all 5 predicted trajectores with their chosen probabilities. For other vehicles, the most likely prediction trajectory is shown instead. Oracle trajectory is shown for vehicle 23 as the oracle outputs are different from top-1 results.}
    \label{fig:trajs}
\end{figure}

To visualize results, Figure~\ref{fig:trajs} compares the model predictions with the ground truth trajectories. The grey and red lines represent the historical and future ground truth trajectories, respectively. The black dots mark the current positions of the vehicles. The light dashed lines represent the predicted trajectories. The light dashed line represents the most-likely (i.e., top-1) predicted trajectory given by the model. One exception is that two vehicles (i.e., vehicle indexed 23 and 27) are selected to visualize all 5 predicted trajectores with their chosen probabilities. The solid dashed lines represent the most likely prediction trajectory either predicted by the model (i.e., top-1) or the oracle (i.e., min@$K$) in terms of FDE in denormalized global coordinates. Results show that the model predictions are generally close to the ground truth trajectories. By normalizing the lateral and longitudinal space, the model can grasp lateral behaviors with decent accuracies. 

For two vehicles with all K trajectories plotted, the text label shows the index of vehicles, trajectories, and probabilities. For example, ``27:4 (0.14)'' means the trajectory is predicted for vehicle 27 for the 5th trajectory (label starts from 0) with a probability of 0.14. By observation, trajectory indexed 3 always represent a ``go straight'' intention whereas other modes represent different lateral behaviors.

Table~\ref{tab:ablation_pedestrian} compares RankFormer against baselines and ablated variants on the ETH/UCY pedestrian trajectory datasets. There are five sub-datasets---ETH, Hotel, Univ, Zara1, and Zara2---each captured at 2.5~fps. Following prior work, 8 timestamps (3.2~s) provide historical context and the subsequent 12 timestamps (4.8~s) are predicted. Ablation columns remove spatial attention while varying~$K$; the unmarked RankFormer column is the full model with spatial attention and $K{=}5$. Results show that in most cases the model choses the ``straight forward'' mode without considering the modes of left/right turns. Unlike the vehicular case, since for most cases the pedestrians densities are low and walk in linear lines, the model's benefits are not revealed.

\begin{table*}[!t]
    \caption{Performance comparison and ablation studies on the ETH/UCY pedestrian trajectory datasets.
    Metrics are reported as ADE/FDE $\downarrow$ (m).
    Panel~(a) reports the most likely trajectory with $K{=}1$.
    Panel~(b) reports minADE/minFDE over the best of $K$ samples ($K{=}20$ for baselines).
    Baseline results for Linear, LSTM, SGAN~\cite{gupta2018social}, and Trajectron++~\cite{salzmann2020trajectron++} are reported for reference.
    Bold entries denote the best result in each row for ADE and FDE, respectively.}
    \label{tab:ablation_pedestrian}
    \centering
    \renewcommand{\arraystretch}{1.12}
    \setlength{\tabcolsep}{2.5pt}
    \footnotesize
    \begin{tabular}{l | cccc | c c c}
    \toprule
    \multicolumn{8}{c}{\textbf{(a) ADE/FDE $\downarrow$ (m),} most likely or only one trajectory prediction} \\
    \midrule
    Dataset
        & Linear
        & LSTM
        & \makecell{S-LSTM}
        & \makecell{Trajectron++}
        & \textbf{RankFormer}$^{\dagger}$
        & \textbf{RankFormer}$^{\ddagger}$
        & \textbf{RankFormer} \\
    \midrule
    ETH     & 1.33/2.94 & 1.09/2.41 & 1.09/2.35 & \textbf{0.71/1.66} & 1.03/1.99 & 1.08/2.14 & 1.07/2.17 \\
    Hotel   & 0.39/0.72 & 0.86/1.91 & 0.79/1.76 & \textbf{0.22/0.46} & 0.33/0.63 & 0.34/0.65 & 0.34/0.64 \\
    Univ    & 0.82/1.59 & 0.61/1.31 & 0.67/1.40 & \textbf{0.44}/1.17 & 0.51/1.13 & 0.51/\textbf{1.12} & 0.53/1.17 \\
    Zara1   & 0.62/1.21 & 0.41/0.88 & 0.47/1.00 & \textbf{0.30/0.79} & 0.44/0.97 & 0.44/0.99 & 0.43/0.95 \\
    Zara2   & 0.77/1.48 & 0.52/1.11 & 0.56/1.17 & \textbf{0.23/0.59} & 0.32/0.70 & 0.32/0.71 & 0.32/0.71 \\
    \midrule
    Average & 0.79/1.59 & 0.70/1.52 & 0.72/1.54 & \textbf{0.38/0.93} & 0.53/1.08 & 0.54/1.12 & 0.54/1.13 \\
    \midrule
    \addlinespace[1pt]
    \midrule
    \multicolumn{8}{c}{\textbf{(b) minADE/minFDE $\downarrow$ (m), Best of $K$ Samples}} \\
    \midrule
    Dataset
        & \makecell{S-GAN}
        & STAR
        & \makecell{Trajectron++}
        & \makecell{AgentFormer}
        & \textbf{RankFormer}$^{\dagger}$
        & \textbf{RankFormer}$^{\ddagger}$
        & \textbf{RankFormer} \\
    \midrule
    ETH     & 0.81/1.52 & 0.36/0.65 & 0.39/0.83 & \textbf{0.26/0.39} & 1.03/1.99 & 0.65/1.15 & 0.71/1.39 \\
    Hotel   & 0.72/1.61 & 0.17/0.36 & 0.12/0.21 & \textbf{0.11/0.14} & 0.33/0.63 & 0.20/0.35 & 0.21/0.38 \\
    Univ    & 0.60/1.26 & 0.31/0.62 & \textbf{0.20/0.44} & 0.26/0.46 & 0.51/1.13 & 0.30/0.63 & 0.31/0.64 \\
    Zara1   & 0.34/0.69 & 0.26/0.55 & \textbf{0.15}/0.33 & \textbf{0.15/0.23} & 0.44/0.97 & 0.27/0.55 & 0.29/0.59 \\
    Zara2   & 0.42/0.84 & 0.22/0.46 & \textbf{0.11}/0.25 & 0.14/\textbf{0.24} & 0.32/0.70 & 0.19/0.38 & 0.20/0.41 \\
    \midrule
    Average & 0.58/1.18 & 0.26/0.53 & 0.19/0.41 & \textbf{0.18/0.29} & 0.53/1.08 & 0.32/0.61 & 0.34/0.68 \\
    \bottomrule
    \end{tabular}

    \vspace{2pt}
    \raggedright
    \footnotesize
    \textit{Note:} $^{\dagger} K{=}1$;
    $^{\ddagger}$ $K{=}6$ without spatial attention;
    unmarked RankFormer is the full model with spatial attention, $K{=}6$.
\end{table*}

\subsection{Model Interpretation}
Since there are Transformer attention layers to process the interactions between vehicles, it is possible to interpret the model's behavior by visualizing the ``flow'' of values within the deep learning model to explore the influence of surrounding vehicles on the vehicle of interest. In practice, just using the attention weights between vehicles is not enough to understand the model's behavior. Thus, we used a normalized value of the attention weights multiplied by the value of the other vehicle's embedding vector to visualize the inter-vehicle impacts on trajectories.

Let $\alpha_{i,j}$ be the attention weight between vehicle $i$ and vehicle $j$. Let $\mathbf{v}_j$ be the embedding vector of vehicle $j$. Then, $I_{i,j} = ||\alpha_{i,j} \cdot \mathbf{v}_j||$ represents the magnitude of the ``impact'' from vehicle $j$ to vehicle $i$. Considering there are $J$ vehicles in the scene at time stamp $t$, the total impact from all vehicles to vehicle $i$ at time stamp $t$ is $I_i(t) = \sum_{j=1}^{J} I_{i,j}(t)$. The normalized impact is $I_i(t) / \sum_{j=1}^{J} I_{i,j}(t)$. In practice, most impacts are from the vehicle itself instead of the surrounding vehicles. Thus, to visualize to impacts, we ignore self impact and only visualize the impacts from other vehicles greater than a threshold (e.g., 0.05).

\begin{figure}[!h]
    \centering
    \includegraphics[width=\linewidth]{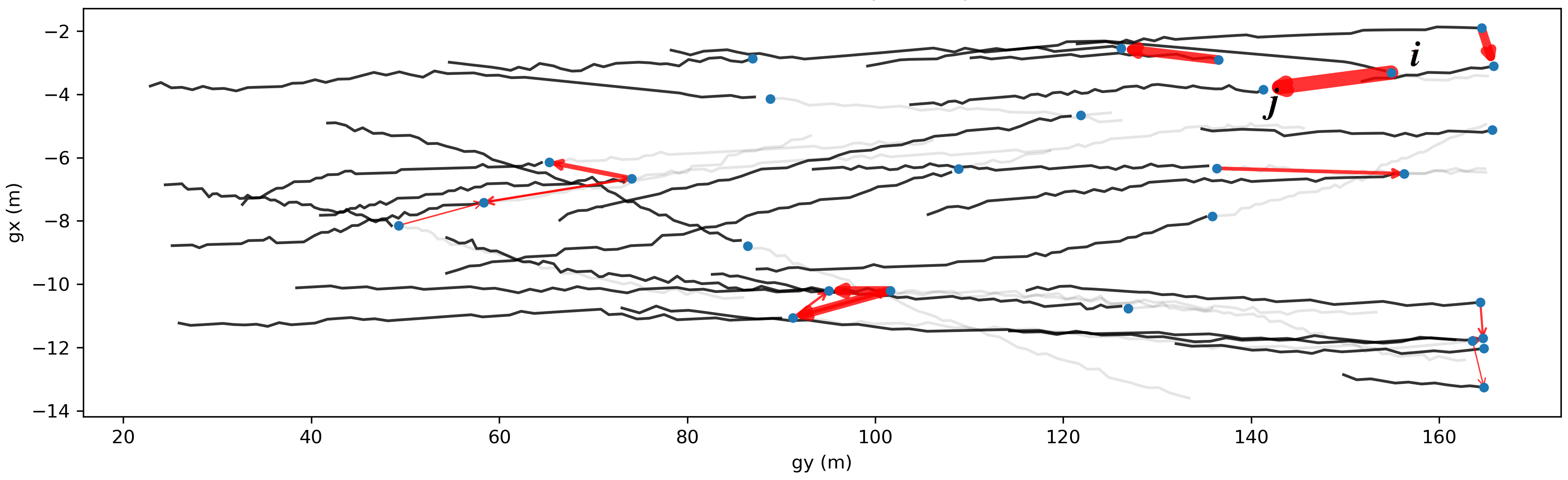}
    \caption{Normalized inter-vehicle impacts ($\sum \|\text{attention weights} \times \text{values}\|$) in a highway scene for the logits of vehicles. Arrow thickness reflects impact magnitude.}
    \label{fig:attn}
\end{figure}

Figure~\ref{fig:attn} shows an example scene in local coordinates ($g_x$, $g_y$). The black and gray lines correspond to the historical and future trajectories, respectively. Blue dots mark each vehicle's current position. Red arrows connect a vehicle of interest to surrounding vehicles whose embedding values are passed through the attention layer; arrow thickness encodes the normalized impact $I_{i,j}$ after discarding self-attention and links below the visualization threshold. Thick arrows therefore indicate strong inter-vehicle influence, while thin arrows indicate weaker but still non-negligible links.

By observing the vehicles, it is clear that the vehicle with cross lane intentions are likely to have higher attention weights to the surrounding vehicles. For example, on the top right corner, vehicle $i$ is a lane-changing vehicle has a decent attention weight to the vehicle $j$ on the back in the right lane. This is similar to the patterns of other cross lane change vehicles, which is similar to the real attention patterns of real-world driver's behavior to focus on the vehicles behind them when changing lanes.



\section{Conclusion}
In this study, we introduced a Transformer-based architecture to predict agent multi-modal trajectories with their probabilities. The model achieved good performance on vehicular dataset, although there is a gap between state-of-the-art models for the pedestrain datasets.

By analysis, several important designs are found useful to enhance model performance. Unlike other models, the structure of the model is elegant with an encoder and two parallel decoders. One decoder focuses on predicting trajectories, the other focuses on modeling agent-agent relationships to ``rank'' the suggested trajectories. The reason for this design is to let the model proposes trajectories for each vehicle independently without interfered by neighboring vehicles. After proposing the trajectories, the second decoder ranks the trajectories based on the agent-agent relationships and generates the final probabilities. The assumption beneath the dual decoder design is that trajectories can be mostly fitted given historical trajectories for short term trajectory prediction.

Besides employing a pure transformer design, we also make some special modificiations for the agent-agent attention module. Learned from traditional graph-based methods, we enhance the spatial encoding module to consider ego-centric velocity and acceleration of neighboring vehicles. All those ego-centric information transitions are applied in a vectorized way to keep the parallel computation of the Transformer layers. Besides the data augmentation for agent-agent relationships, we also apply an inter-modal Transformer block to let the information from different candidate trajectories to communicate each other before generating the final trajectory probabilites. Combining all these improvements, we identify an efficient architecture to predict multi-modal agent trajectory predictor with probabilities.

Although the study finds some tricks to improve agent multi-modal trajectory predictions, there are still some limitations to be addressed in the future. First, the most cannot consistently outcompete the constant speed results in the pedestrian dataset since most pedestrian follows a constant speed walking strategy. Second, compared with CVAE methods, the memory consumption is higher limited by aligning different modes along a new tensor dimension. Thus, the study only fits $K=5$ different modes as the complexity of the model also increases in a quadratic way as $K$ increases. In practice, if there are more than 6 modes of intentions, many modes may collapses to useless values with near zero probabilities and could generate drifted trajectories. To give initially inaccurate modes opportunities to learn, we apply epsilon-greedy exploration to the regression assignment early in training, while supervising the ranker independently with distance-based soft targets. For future studies, researchers should apply a more efficient method to learn latent space to generate more modes of intentions.

Another limitation comes from the data sources. Two evaluated datasets are lack of contextual information such as lanes, border, and obstacles. Thus, the current version of the model doesn't encode the context-aware environments. Besides, there is not heterogeneous agent types within one scene to train more generous models. For future studies, the model should incorporate more contextual information into the model to improve the performance. Furthermore, given raw video inputs, it is likely to design an end-to-end model to directly generate predictions given running inputs.

\section*{Acknowledgment}
The authors would like to thank the Ubiquitous Traffic Eye team for providing open source vehicle trajectory data by Southeast University, China.

\section*{Conflict of Interest}
The authors declare no competing financial interests or personal relationships relevant to this work.

\ifCLASSOPTIONcaptionsoff
  \newpage
\fi

\bibliographystyle{IEEEtran}
\bibliography{bibtex/bib/traj}

\end{document}